\documentclass[conference]{IEEEtran}
\IEEEoverridecommandlockouts
\usepackage{cite}
\usepackage{amsmath,amssymb,amsfonts}
\usepackage{algorithmic}
\usepackage{graphicx}
\usepackage{textcomp}
\usepackage{xcolor}
\def\BibTeX{{\rm B\kern-.05em{\sc i\kern-.025em b}\kern-.08em
    T\kern-.1667em\lower.7ex\hbox{E}\kern-.125emX}}

\usepackage{placeins}
\usepackage[hyphens]{url}
\usepackage{hyperref}
\hypersetup{colorlinks=true,breaklinks=true}

\usepackage{mathrsfs,amssymb,amsmath,amsthm}

\usepackage{mathtools} 

\usepackage{tabularx} 
\usepackage{booktabs}
\usepackage[linesnumbered,ruled,vlined]{algorithm2e}
\SetKwInOut{Parameter}{Parameter}
\usepackage{xargs}                      
\usepackage[colorinlistoftodos,prependcaption,textsize=tiny]{todonotes}
\newcommandx{\unsure}[2][1=]{\todo[linecolor=red,backgroundcolor=red!25,bordercolor=red,#1]{#2}}
\newcommandx{\change}[2][1=]{\todo[linecolor=blue,backgroundcolor=blue!25,bordercolor=blue,#1]{#2}}
\newcommandx{\info}[2][1=]{\todo[linecolor=OliveGreen,backgroundcolor=OliveGreen!25,bordercolor=OliveGreen,#1]{#2}}
\newcommandx{\improvement}[2][1=]{\todo[linecolor=Plum,backgroundcolor=Plum!25,bordercolor=Plum,#1]{#2}}
\newcommandx{\thiswillnotshow}[2][1=]{\todo[disable,#1]{#2}}
\newcommand{\iu}{{i\mkern1mu}}
\newcommand{\norm}[1]{\left\lVert#1\right\rVert}
\SetKwInput{KwInput}{Input}                
\SetKwInput{KwOutput}{Output}              

\usepackage{caption}
\usepackage{bbm}
\usepackage{subcaption}
\usepackage[detect-all]{siunitx}
\usepackage{cleveref}
\newcommand\copyrighttext{%
	\footnotesize \copyright~2023 IEEE. Personal use of this material is permitted. Permission from IEEE must be obtained for all other uses, in any current or future media, including reprinting/republishing this material for advertising or promotional purposes,creating new collective works, for resale or redistribution to servers or lists, or reuse of any copyrighted component of this work in other works.}
\newcommand\copyrightnotice{%
	\begin{tikzpicture}[remember picture,overlay]
		\node[anchor=south,yshift=10pt] at (current page.south) {\fbox{\parbox{\dimexpr\textwidth-\fboxsep-\fboxrule\relax}{\copyrighttext}}};
	\end{tikzpicture}%
}

\begin{document}

\title{Spectral Batch Normalization: Normalization in the Frequency Domain\\}

\author{\IEEEauthorblockN{Rinor Cakaj}
	\IEEEauthorblockA{\textit{Image Processing} \\
		\textit{Robert Bosch GmbH \& University of Stuttgart}\\
		71229 Leonberg, Germany \\
		Rinor.Cakaj@de.bosch.com}
	\and
	\IEEEauthorblockN{Jens Mehnert}
	\IEEEauthorblockA{\textit{Image Processing} \\
		\textit{Robert Bosch GmbH}\\
		71229 Leonberg, Germany \\
		JensEricMarkus.Mehnert@de.bosch.com}
	\and
	\IEEEauthorblockN{Bin Yang}
	\IEEEauthorblockA{\textit{ISS} \\
		\textit{University of Stuttgart}\\
		70550 Stuttgart, Germany \\
		bin.yang@iss.uni-stuttgart.de}
}

\maketitle

\begin{abstract}
Regularization is a set of techniques that are used to improve the
generalization ability of deep neural networks. In this paper, we introduce
\emph{spectral batch normalization} (SBN), a novel effective method to improve
generalization by normalizing feature maps in the frequency (spectral) domain.
The activations of residual networks without batch normalization (BN) tend to
explode exponentially in the depth of the network at initialization. This leads
to extremely large feature map norms even though the parameters are relatively
small. These explosive dynamics can be very detrimental to learning.
BN makes weight decay regularization on the scaling factors $\gamma, \beta$
approximately equivalent to an additive penalty on the norm of the feature maps,
which prevents extremely large feature map norms to a certain degree. It was
previously shown that preventing explosive growth at the final layer at
initialization and during training in ResNets can recover a large part of Batch
Normalization's generalization boost.
However, we show experimentally that, despite the approximate additive penalty
of BN, feature maps in deep neural networks (DNNs) tend to explode at the
beginning of the network and that feature maps of DNNs contain large values
during the whole training. This phenomenon also occurs in a weakened form in
non-residual networks. Intuitively, it is not preferred to have large values in
feature maps since they have too much influence on the prediction in contrast to
other parts of the feature map. 
SBN addresses large feature maps by normalizing them in the
frequency domain. In our experiments, we empirically show that SBN prevents
exploding feature maps at initialization and large feature map values during the
training. Moreover, the normalization of feature maps in the frequency domain
leads to more uniform distributed frequency components. This discourages the
DNNs to rely on single frequency components of feature maps. These, together
with other effects (e.g. noise injection, scaling and shifting of the feature
map) of SBN, have a regularizing effect on the training of residual and
non-residual networks.
We show experimentally that using SBN in addition to standard regularization
methods improves the performance of DNNs by a relevant margin, e.g. ResNet50
on CIFAR-100 by 2.31\%, on ImageNet by 0.71\% (from 76.80\% to 77.51\%) and VGG19 on CIFAR-100 by 0.66\%.
\end{abstract}
\copyrightnotice
\section{Introduction} \label{Introduction}

Deep neural networks contain multiple non-linear hidden layers which make them
powerful machine learning systems \cite{2014_Srivastava}. However, such networks
are prone to overfitting \cite{2013_Wan_CONF} due to the
limited size of training data or the high capacity of the model.
Overfitting describes the phenomenon where a neural network
(NN) perfectly fits the training data while achieving poor performance on the
test data. Regularization is a set of techniques used to reduce overfitting and
is therefore a key element in deep learning \cite{2016_Goodfellow_BOOK}. It
allows the model to generalize well to unseen data. 

Many methods have been developed to regularize DNNs, weight penalties as $L_1$-regularization \cite{1996_Tibshirani}
and weight decay \cite{1991_Krogh_CONF}, soft weight sharing \cite{1992_Nowlan},
dropout \cite{2014_Srivastava}, data augmentation \cite{2019_Shorten} and
ensemble learning methods \cite{1999_Opitz}. 

Normalization techniques \cite{2015_Ioffe_CONF, 2016_Ba, 2020_Wu, 2016_Ulyanov}
like batch normalization (BN) \cite{2015_Ioffe_CONF} normalize features by
subtracting the mean and dividing by the standard deviation computed across
different dimensions of a feature map. In some cases such normalization
techniques act as regularizers, eliminating the need for dropout
\cite{2015_Ioffe_CONF}. There are different explanations for the regularizing
effects of BN: (i) the stochastic uncertainty of the batch statistics can
benefit generalization \cite{2020_Wu}, (ii) BN reduces the explosive growth of
feature maps in deeper layers which acts as a regularizer
\cite{2021_Dauphin_CONF} and (iii) BN discourages reliance on a single neuron
and encourages different neurons to have equal magnitude in the
sense that corrupting individual neurons does not harm generalization
\cite{2019_Luo_CONF}.

The feature maps (i.e. activations) of residual networks without BN tend to
explode exponentially in the depth of the network at initialization
\cite{2019_Zhang_CONFa}. This leads to
extremely large feature map norms even though the parameters are relatively
small \cite{2021_Dauphin_CONF}. Large and exploding feature maps also occur in a
weakened form in non-residual networks. These explosive dynamics can be very
detrimental to learning \cite{2019_Zhang_CONFa}. 

Intuitively, it is not preferred to have large values in the feature maps since
they have too much influence on the prediction in contrast to other parts of the
feature map. Therefore some sort of scaling is needed to restrict the influence
of single outputs and to distribute the decision making process on
a larger part of the feature map. 

With ormalization techniques is weight decay regularization on the scaling and
shifting factors $\gamma, \beta$ approximately equivalent to an additive penalty
on the norm of the feature maps \cite{2021_Dauphin_CONF}. This prevents
exploding features during training to some degree. Dauphin et al.
\cite{2021_Dauphin_CONF} showed experimentally that preventing explosive growth
at the final layer at initialization and during training can recover a large
part of BNs generalization boost. 

\begin{figure*}[tb]
	\centering
	\begin{subfigure}{0.49\linewidth}
		\centering
		\includegraphics[width=\columnwidth]{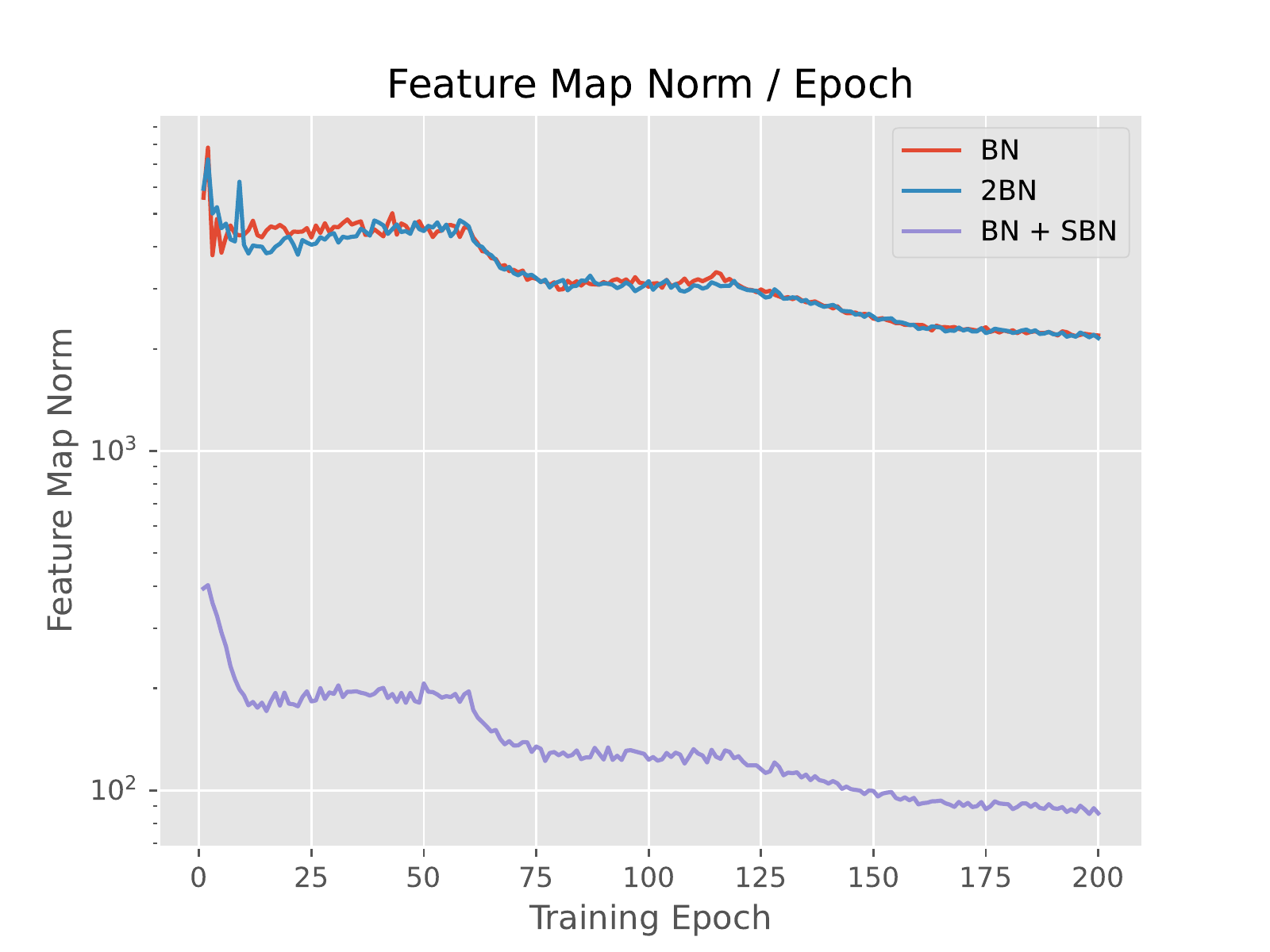}
		\subcaption{Feature map norm of BN, 2BN and BN + SBN layer per epoch for an arbitrary batch in ResNet50 trained on CIFAR$100$.}
		\label{fig:resnet50_cifar100_30_01}
	\end{subfigure}\hfill
	\begin{subfigure}{0.49\linewidth}
		\centering
		\includegraphics[width=\columnwidth]{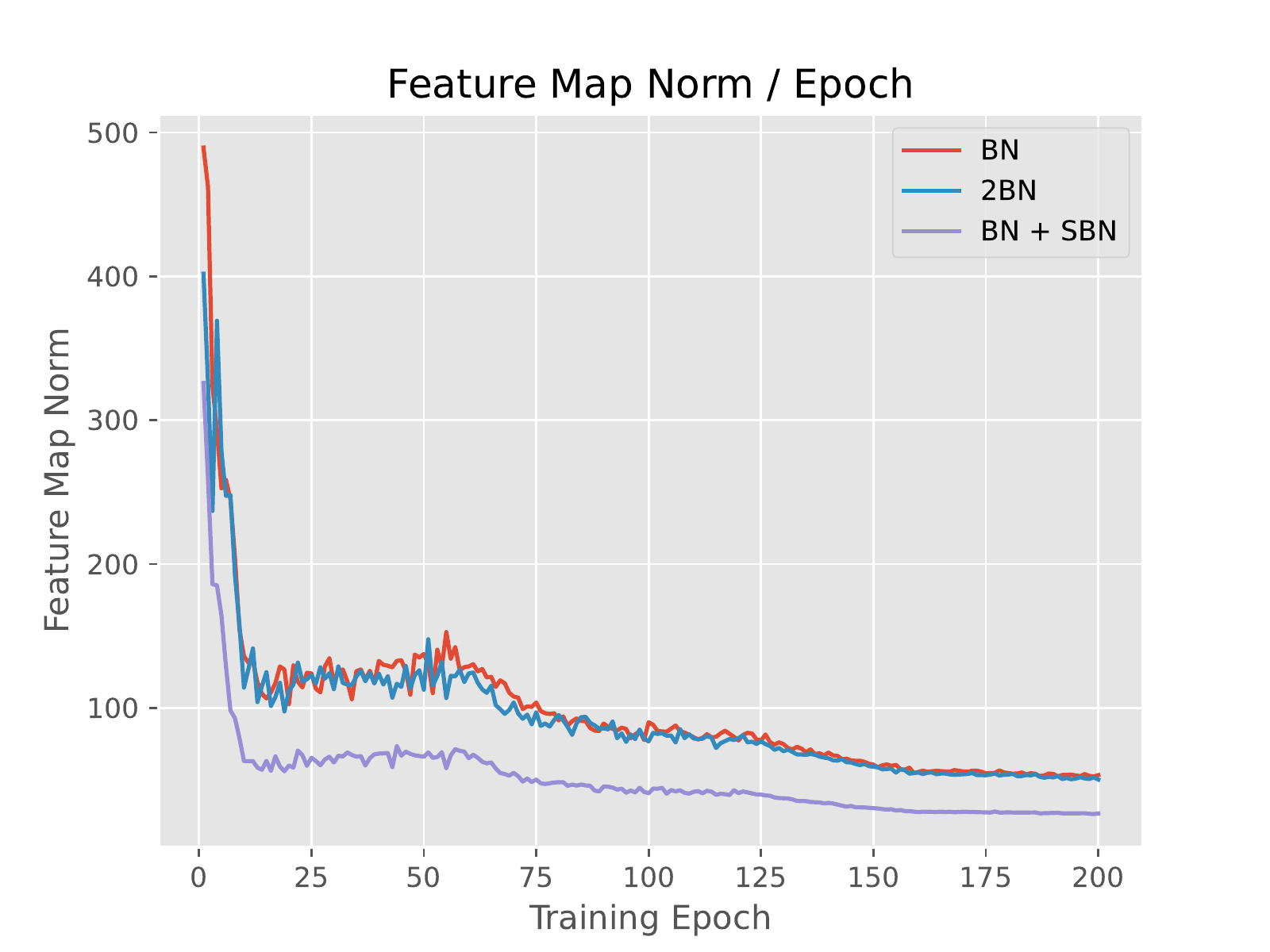}
		\subcaption{Feature map norm of BN, 2BN and BN + SBN layer per epoch for an arbitrary batch in VGG19 trained on CIFAR$100$.}
		\label{fig:vgg_cifar100_conv1_2}
	\end{subfigure}
	\caption{Feature map norms per epoch. The subfigures show that using SBN leads to smaller feature map norms. Note that the Subfigure \ref{fig:resnet50_cifar100_30_01} has
		logarithmic scaling which makes the explosion not so visible. } \label{fig:feature_map_norm}
\end{figure*}

However, in our analysis of feature maps for different networks with BN we show
that feature maps tend to explode at the beginning of the training and that
they contain large values during the whole training despite the
additive penalties $\gamma, \beta$. Figure \ref{fig:feature_map_norm} compares
the feature map norm over the course of training of a BN layer (``BN''), two
subsequent BN layers (``2BN'') and a \emph{Spectral Batch Normalization} 
(SBN) layer following a BN layer (``BN + SBN'') in a ResNet50 and VGG$19$
trained on CIFAR-$100$ for an arbitrary batch. The figure shows that SBN
prevents exploding feature maps at initialization and large feature map norms
during the whole training. Note that the Subfigure
\ref{fig:resnet50_cifar100_30_01} has logarithmic scaling. 

To prevent exploding feature maps at initialization and large feature maps
during the whole training, SBN normalizes the feature maps in the frequency
(spectral) domain.

Firstly, SBN computes the 2-dimensional discrete Fourier transform (DFT) of the
feature maps. Secondly, it computes the channel-wise mean and standard deviation
of the DFT coefficients across all frequencies and the mini-batch. Thirdly, the
frequency components are normalized. After a re-scaling and re-shifting by
learned parameters, the feature map is transformed back into the spatial domain
using the inverse DFT. 

SBN prevents exploding feature maps at initialization and large values in
feature maps during the whole training which would destabilize the training and
would lead to a more complex model compared to DNNs with small feature maps.
Furthermore, SBN has more positive effects: (i) it inserts stochastic
noise in the frequency domain which makes the network more robust to small
perturbations in the frequency domain and (ii) re-scaling and re-shifting gives
the network the opportunity to weight channels in the frequency domain and (iii)
it leads to more uniform distributed frequency components in the feature maps
which discourage the network to rely on single frequency components.

In Section \ref{Experiments} we will show experimentally that using SBN in
addition to BN prevents large values in the feature maps during the whole
training and therefore reduces overfitting in DNNs. The feature maps are scaled
down to smaller values in comparison to only using BN. Furthermore, we will show
that SBN increases the accuracy of networks by a relevant margin compared to
the original models. 

\subsection{Contributions of this Paper and Applications}

In this work, we present \emph{spectral batch normalization}, a novel effective
method to improve generalization by normalizing feature maps in the frequency
(spectral) domain. Our core contributions are:

\begin{itemize}
	\item Introducing spectral batch normalization.
	\item Analyzing the impact of our method on the weights and feature maps (in the spatial and frequency domain) during the training process.
	\item Showing experimentally that using spectral batch normalization in
addition to standard regularization methods increases the performance of various
different network architectures on CIFAR-$10$/CIFAR-$100$ and on ImageNet. The
additional gains in performance of ResNet$50$ is on CIFAR-$10$ by $1.40\%$, on
CIFAR-$100$ by $2.32\%$, on TinyImageNet by $2.36\%$ and on ImageNet by $0.71\%$
are worth noting. Moreover, the performance of non-residual networks is also
improved, e.g. VGG$19$ on CIFAR-$100$ by $0.66\%$.
\end{itemize}

\section{Related Work} \label{Related-Work}

\subsection{Regularization}

Regularization is one of the key elements of deep learning
\cite{2016_Goodfellow_BOOK}, allowing the model to generalize well to unseen
data even when trained on a finite training set or with an imperfect
optimization procedure \cite{2017_Kukacka}. There are several techniques to
regularize NNs which can be categorized into groups. Data augmentation
methods like cropping, flipping, and adjusting brightness or sharpness
\cite{2019_Shorten} and cutout \cite{2017_DeVries}
transform the training dataset to avoid overfitting.
Regularization techniques like dropout \cite{2014_Srivastava},
dropblock \cite{2018_Ghiasi_CONF}, dropconnect
\cite{2013_Wan_CONF} drop neurons or weights from the NN during training to
prevent units from co-adapting too much \cite{2014_Srivastava}. Furthermore, NNs
can be regularized using penalty terms in the loss function. Weight decay
\cite{1991_Krogh_CONF,	2019_Loshchilov_CONF} encourages the weights of the NN to be small in magnitude. The $L_1$-regularization \cite{1996_Tibshirani} forces
the weights of non-relevant features to zero. 

\subsection{Normalization}

Normalization methods are an essential building block of the most successful
deep learning architectures and have enabled training very deep residual
networks \cite{2019_Zhang_CONFa}. 

Batch normalization (BN) \cite{2015_Ioffe_CONF} normalizes features by
subtracting the mean and dividing by the standard deviation computed within a
mini-batch. The normalization is done to reduce the internal covariate shift,
i.e. the change in the distribution of network activations due to the change in
network parameters during training. The features are then scaled and shifted by
learned parameters to maintain the expressive power of the network
\cite{2016_Goodfellow_BOOK}. During the training, a moving average of the means
and variances is computed which are then used during inference. Batch
normalization enables faster and more stable training of deep NNs. Moreover, it
also acts as a regularizer, in some cases eliminating the need for dropout. 


The effect of batch normalization is dependent on the mini-batch size and it is
not obvious how to apply it to recurrent neural networks. Layer normalization
\cite{2016_Ba} computes the mean and variance used for normalization from all
the features in a layer on a single training case. It improves the training time
and generalization performance of several sequential models (RNN
\cite{Rumelhart1987}).


Instance normalization \cite{2016_Ulyanov} normalizes across the width and
height of a single feature map of a single example, i.e. not across the channels
like layer normalization. It works well for generative models.


Layer normalization and instance normalization have limited success in visual
recognition. Therefore, Wu et al. \cite{2020_Wu} present group normalization
which avoids normalizing along the batch dimension but divides the channels into
groups and computes the mean and variance within each group for normalization.


Weight normalization \cite{2016_Salimans_CONF} is a weight reparameterization
approach that accelerates the convergence of SGD optimization. It
reparameterizes the weight vectors of each layer such that the length of those
weight vectors is decoupled from their direction. In detail, they express the
weight vector $w \in \mathbb{R}^d$ by $w = \Psi(v, g) = \frac{g}{\norm{v}} v$
where $v \in \mathbb{R}^d$ is the new weight vector and $g \in \mathbb{R}$ the
scalar parameter.

There is a number of works that analyzed the reasons for the success of
normalization methods. In the case of BN, Ioffe et al. \cite{2015_Ioffe_CONF}
stated that it reduces the internal covariate shift, i.e. the change in the
distribution of each layer's input during training. However, Santurkar et al.
\cite{2018_Santurkar_CONF} demonstrate that the distributional stability of
layer inputs has little to do with the success of BN. Instead, they point
out that BN makes the optimization landscape significantly smoother. This would
induce a more predictive and stable behavior of the gradients, allowing faster
training. It has been shown that the activations of residual networks without BN
tend to explode exponentially in the depth of the network initialization
\cite{2019_Zhang_CONFa}. Preventing the explosive growth
at the final layer at initialization and during the training can recover a large
part of BNs generalization effect \cite{2021_Dauphin_CONF}.

\subsection{Frequency}


Due to the dual of the convolution theorem, stating that multiplication in the
time/spatial domain is equivalent to convolution in the frequency domain, the
DFT is used in deep learning to provide a significant speedup in the computation
of convolutions. Moreover, Rippel et al. \cite{2015_Rippel_CONF} showed that the
frequency domain also provides a powerful representation to model and train
CNNs. They introduced spectral pooling which performs a dimensionality
reduction by truncating the representation in the frequency domain. 


The DFT is also used to regularize DNNs. Spectral dropout \cite{2019_Khan}
prevents overfitting by eliminating weak Fourier domain coefficients below a
fixed threshold and randomly dropping a fixed percentage of the remaining
Fourier domain coefficients of the neural network activations. 


Furthermore, the frequency domain can be used to prune NNs \cite{2018_Liu_CONF}.

\section{Method} \label{Method}

In order to counteract exploding or large feature maps, we want to use a
frequency decomposition. It allows manipulating an input across its various
length-scales of variation, and as such provides a natural framework for the
manipulation of data with spatial coherence \cite{2015_Rippel_CONF}. 

Inspired by batch normalization \cite{2015_Ioffe_CONF}, we introduce
\emph{spectral batch normalization}, which has a regularizing effect on DNNs, by
normalizing feature maps in the frequency domain. SBN prevents explosive
dynamics during the initialization and leads to smaller feature maps during the
whole training. To introduce spectral batch normalization, we need the discrete
Fourier transformation.

\subsection{The Discrete Fourier Transform}

The discrete Fourier transform converts an input array of real (or
complex) numbers into another sequence of complex numbers. In our case, it is a
frequency domain representation of the original spatial input sequence. The 2D
DFT of a matrix $\mathbf{x} \in \mathbb{R}^{H \times W}$ is defined by
\begin{align}
	\hat{\mathbf{x}}_{k,l} \coloneqq \mathscr{F}(\mathbf{x})_{k,l} \coloneqq \sum_{h=0}^{H-1}
	\sum_{w=0}^{W-1} \mathbf{x}_{h,w} \cdot e^{-2 \pi \iu \cdot (\frac{h \cdot k}{H}
		+ \frac{w \cdot l}{W})}
\end{align}
for $k=0, \dots, H-1$ and $l=0, \dots, W-1$. Its inverse transform is given by $\mathscr{F}^{-1}(\cdot) = \mathscr{F}^*(\cdot) / (HW)$, i.e. the conjugate of the transform normalized by $1/(HW)$.

Intuitively, the DFT decomposes an input array into the frequencies contained
in the input sequence. Roughly speaking, it compares a basis of complex
sinusoidal functions with the input sequence and computes the similarity between
the complex sinusoidal functions and the input sequence. 

The DFT of a real signal is Hermitian-symmetric, i.e. $\mathbf{\hat{x}}_{k,l} =
\text{conj}(\mathbf{\hat{x}}_{H-i,W-j})$. Therefore, only $HW/2$
complex numbers are needed to represent the real signal in the frequency domain.
This does not reduce the effective dimensionality of the transformed input
since each DFT coefficient consists of a real and imaginary component. 


Before we explain spectral batch normalization, we have to discuss how to
propagate the gradient through a Fourier transform layer. This is well described
by Rippel et al. \cite{2015_Rippel_CONF}. Let $\mathbf{x} \in
\mathbb{R}^{H \times W}$ be the input and $\mathscr{F}(x) = \mathbf{y} \in
\mathbb{C}^{H \times W}$ be the output of a DFT. Moreover, let $Z \colon
\mathbb{C}^{H \times W} \to \mathbb{R}$ be a real-valued loss function applied
to $\mathbf{y}$ which can be considered as the remainder of the forward pass.
The DFT is a linear operator, therefore its gradient is the transformation
matrix itself. During back-propagation, the gradient is conjugated
\cite{2021_Bassey}. This corresponds to the application of the inverse transform
\begin{align}
	\frac{\partial Z}{\partial \mathbf{x}}  = \mathscr{F}^{-1} \left(\frac{\partial
		Z}{\partial y}\right).
\end{align}

We use the real 2D DFT implementation of PyTorch \cite{2019_Paszke_CONF}, which
uses the FFT algorithm to compute the transformation. The implementation of the
inverse real 2D DFT in PyTorch uses zero-padding to get the original array size
$H \times W$ for signals with odd length in a transformed dimension.

\subsection{Spectral Batch Normalization} \label{SBN}

Let $\mathbf{X} \in \mathbb{R}^{B \times C \times H \times W}$ be a feature map.
The introduced spectral batch normalization block first computes the real
2-dimensional discrete Fourier transform $\mathbf{\hat{X}} \in \mathbb{C}^{B
	\times C \times H \times W/2}$ of the last two
dimensions of the feature map $\mathbf{X}$. Secondly, it computes the
channel-wise mean and standard deviation of the DFT coefficients across all
frequencies and a mini-batch. Hence, the resulting mean and
variance have size $1 \times C \times 1 \times 1$.  During the training, a
moving average of the means and standard deviations is computed which are then
used during inference. Then, the transformed feature maps are normalized
channel-wise using the computed mean and standard deviation (like in batch
normalization \cite{2015_Ioffe_CONF}). To recover the representation power of
the layer, the feature maps are scaled and shifted channel-wise by learnable
parameters $\gamma \in \mathbb{R}^C$ and $\beta \in \mathbb{R}^C$. After the
normalization and scale/shift step, the feature map is transformed back using
the inverse real DFT. SBN is presented in Algorithm \ref{algorith_SN}. 

\begin{algorithm}[!ht] 
	\DontPrintSemicolon
	\Parameter{momentum=$\lambda \in (0,1)$, epsilon=$\epsilon \in \mathbb{R}$, running\_mean=$\mu_r \in \mathbb{R}^C = \mathbf{0}$, running\_variance=$\sigma_r^2 \in \mathbb{R}^C = \mathbf{1}$}
	\KwInput{$\mathbf{x} \in \mathbb{R}^{B \times C \times H \times W}$}
	\KwOutput{$\mathbf{y} \in \mathbb{R}^{B \times C \times H \times W}$}
	
	\tcc{Compute the 2D-DFT of $\mathbf{x}$, i.e. on the last two dimensions.}
	$\mathbf{\hat{x}} = \mathscr{F}(x)$
	
	\tcp{Compute mini-batch mean/variance}
	$\mu = \text{mean}(\mathbf{\hat{x}}, \text{dim}=[0,2,3])$
	
	$\sigma^2 = \text{var}(\mathbf{\hat{x}}, \text{dim}=[0,2,3])$
	
	\tcp{Compute running mean/variance}
	
	$\mu_r = \lambda \cdot \mu + (1- \lambda) \cdot \mu_r$
	
	\tcp{Compute running unbiased variance, n=CHW/2}
	
	$\sigma_r^2 = \lambda \cdot \sigma^2 \cdot \frac{n}{n-1} + (1 - \lambda) \cdot \sigma_r^2$
	
	$\mathbf{\bar{x}} = \frac{\mathbf{\hat{x}} - \mu}{\sqrt{\sigma^2 + \epsilon}}$	\tcp{Normalize}
		
	$\mathbf{\hat{y}} = \gamma \cdot \mathbf{\bar{x}} + \beta$ \tcp{Scale and shift channel-wise}
	
	$\mathbf{y} = \mathscr{F}^{-1}(\mathbf{\hat{y}})$ \tcp{Compute inverse 2D DFT}
	
	\caption{Spectral Batch Normalization Layer} \label{algorith_SN}
\end{algorithm}

At this point, the question could arise why we compute the \textbf{channel-wise}
mean and standard deviation of the feature maps across a mini-batch, i.e. the
resulting mean and variance has size $1 \times C \times 1 \times 1$. 

Different channels in a feature map are often independent and represent
different information depending on the input. Therefore, it is not beneficial to
normalize across different channels. Moreover, computing the mean and
standard deviation across a mini-batch and the channels and \emph{not}
over the frequency components leads for some datasets (e.g. ImageNet) to a huge
number of additional scaling and shifting weights (i.e. $\# H \cdot W$).

We now have a short look at the effects of SBN on the frequency
components in the feature map. The resulting values of a DFT
are complex numbers. Let $z = a + ib$ be a complex number. Every
nonzero complex number can be written in the form $re^{i \theta}$,
where $r \coloneqq \sqrt{a^2 + b^2}$ is the \textbf{magnitude} of $z$, and
$\theta \in [0, 2 \pi)$ is the \textbf{phase, angle} or \textbf{argument} of
$z$. The magnitude of $z$ represents the amplitude of the corresponding
frequency. Hence, the magnitude gives us information about which frequency
components are mainly represented in our feature map. The phase of $z$
represents the phase of the corresponding frequency, i.e. the spatial delay for
that frequency in the feature map. Due to the normalization and
scaling/shifting, the magnitude and the phase of the frequency components are
affected by SBN.

SBN computes the mean and variance of the complex numbers using
\begin{align}
	\hat{\mu} &= \frac{1}{N} \sum_{j=1}^N z_j = \frac{1}{N} \sum_{j=1}^N a_j + i \cdot \frac{1}{N} \sum_{j=1}^N b_j  \label{mu} \\
	\hat{\sigma}^2 &= \frac{1}{N} \sum_{j=1}^N |\hat{z}_j - \hat{\mu}|^2.
\end{align}
Then the mean is substracted from the DFT coefficients. The real and complex
part of the DFT coefficients are changed separately to have zero mean (see
Equation \eqref{mu}). This leads to DFT coefficients which are distributed
around $z_0 = 0 + i \cdot 0$. Subsequently, the DFT coefficients are divided by the standard deviation. 

These steps make the magnitudes of the frequency components smaller which then
leads to smaller values in the feature maps in the spatial domain. Moreover, the
frequency components become more uniformly distributed, i.e. the importance of
frequency components with large magnitudes is reduced. On the other hand, the
frequency components with low magnitudes get a higher influence on the feature
map. The effects on the phase are difficult and cannot be stated clearly. 
The effects of SBN on the the frequency components are discussed
empirically in Section \ref{effects_spectral_domain}.

Lastly, we want to point out where to insert a SBN layer. Empirically, it is
preferred to insert the SBN layer in deeper layers after the BN layer. More
details are discussed in Section \ref{where_SBN}.

\subsection{Regularizing effects}

There are several components of SBN which act as a regularizer during training. 

\subsubsection{Preventing large feature map norm}

As stated in the introduction, the activations of residual networks without BN
tend to explode exponentially in the depth of the network at initialization
\cite{2019_Zhang_CONFa} which can be
detrimental to learning.

Batch normalization prevents exploding feature maps during the training to a
certain degree by normalizing the feature maps. Dauphin et al.
\cite{2021_Dauphin_CONF} showed that most of the regularizing effect of BN comes
from the prevention of explosive growth from cascading to the output during
training. 

Our analysis of feature maps for different networks (with and without residual
connections) trained with BN on different datasets showed that large values in
feature maps occur during the whole training despite the usage of BN.

Using SBN in addition to BN reduces the explosive growth of feature
maps further due to the normalization process in the frequency domain. The
feature maps are scaled down to smaller values compared to only using
BN (see Section \ref{Experiments}).

\subsubsection{Noise Injection}

SBN subtracts a random value (the mean of the mini-batch across each feature map
and mini-batch) from each DFT transformed feature map. Moreover, SBN divides
each DFT transformed feature map by a random value (the standard deviation of
the mini-batch across each feature map and mini-batch) at each step of training.
Because different examples are randomly chosen for inclusion in the mini-batch
at each step, the standard deviation and mean randomly fluctuate.
Both of the sources of noise manipulate the feature maps in the frequency
domain. Hence every layer has to learn to be more robust to a lot of variation
in its inputs \cite{2016_Goodfellow_BOOK}. Therefore the stochastic
uncertainty of the batch statistics acts as a regularizer during training
\cite{2016_Ba, 2020_Wu}. 

\subsubsection{Re-scaling and Re-shifting}

As in the standard BN, we introduce weights $\gamma$ and biases $\beta$ to
recover the representation power of feature maps in the frequency domain by
multiplying the weights and biases with the DFT coefficient. Since weights and
biases are learned through SGD, the network learns to prioritize the channels of
the feature maps in the frequency domain.

\subsubsection{More uniform distribution of frequency components}

As stated in Section \ref{SBN}, SBN leads to more uniformly distributed
frequency components. Hence, the influence of strong frequency components
characterized via high magnitudes is reduced. On the other hand, the importance
of weak frequency components is relatively increased. The effect has a
regularizing effect on the training, since it discourages the DNN to rely on
single frequency components. This is the key difference between SBN and BN,
because BN normalizes feature maps only in the spatial domain. 

\section{Experiments} \label{Experiments}

We experimentally validate the usefulness of our method in supervised image
recognition. We compare DNNs using SBN in addition
to standard regularization methods against DNNs using only standard
regularization methods.

ResNets consist of four modules made up of basic
and “bottleneck” building blocks, a convolutional layer at the beginning of the
network and a linear layer at the end of the network. VGGs have a similar module
structure. Through our experiments, we figured out that SBN should preferably be
inserted in deeper layers of DNNs. The additions to the names of the models
describe in which modules SBN is applied. For example ''ResNet$50$ + SBN$34$``
means that, a ResNet$50$ network is used where in the third and fourth module a
SBN layer is inserted after each BN layer.

In order to achieve a fair comparison, we also added results of ResNets with two
BN layers in the same modules where SBN is inserted, e.g. ``ResNet$50$ +
$2$BN$34$'' is the abbreviation for a ResNet$50$ with two BN layers, one after
the other, in the third and fourth module.

\subsection{Image Classification on CIFAR-$10$/$100$}

We evaluate the performance of SBN on the CIFAR-$10$/$100$ classification
datasets \cite{2009_Krizhevsky}. The CIFAR-$10$/$100$
datasets consist of 50,000 training and 10,000 test $32\times32$ color images.
We used various different network architectures, the CNNs ResNet$18$,
ResNet$34$, ResNet$50$ \cite{2016_He_CONF} and VGG$16$/$19$-BN
\cite{2015_Simonyan_CONF}. 

The experiments were run five times with different random seeds for $200$
epochs, resulting in different network initializations, data orders and
additionally in different data augmentations. For every case we report the mean
test accuracy and standard deviation. We used a $9/1$-split between training
examples and validation examples and saved the best model on the validation set.
This model was then used for evaluation on the test dataset.

The baseline networks ResNet$18$, ResNet$34$ and ResNet$50$ were trained with
data augmentation, weight decay and early stopping. The baseline network
VGG$16$-BN and VGG$19$-BN were trained with data augmentation, weight decay,
early stopping and dropout in the fully connected layers. 

\subsubsection{Implementation Details} 

For our experiments we used PyTorch 1.10.1 \cite{2019_Paszke_CONF} and one
Nvidia GeForce 1080Ti GPU. 

All networks were trained with a batch size of $128$. We used the SGD optimizer
with momentum $0.9$ and initial learning rate $0.1$. For ResNet$18$, ResNet$34$
and ResNet$50$ trained on CIFAR-$10$ we decayed the learning rate by $0.1$ at
epoch $90$ and $136$ and used weight decay with the factor $1e-4$ (as in
\cite{2016_He_CONF}). For all networks trained on CIFAR-$100$ and for
VGG$16$/$19$ with BN trained on CIFAR-$10$, we decayed the learning rate by
$0.2$ at epochs $60$, $120$ and $160$ and used weight decay with the factor
$5e-4$ as in \cite{2016_Zagoruyko_CONF}. Dropout is used in the fully connected
layers in VGG16-BN and VGG19-BN with a dropout rate of $0.5$. We want to point
out that we have not done a hyperparameter search for the learning rate, weight
decay, etc. We used the same setting as in
\cite{2016_He_CONF,2016_Zagoruyko_CONF}. 

The training data is augmented by using random crop with
size $32$ and padding $4$, random horizontal flip and normalization
\cite{2019_Shorten}. The test set is only normalized. The ResNet networks are
initialized Kaiming-uniform \cite{2015_He_CONF}. In the VGG networks the
linear layers are initialized Kaiming-uniform. The convolutional layers are
initialized with a Gaussian normal distribution with mean $0$ and standard
deviation $2/n$ where $n$ is the size of the kernel multiplied with the
number of output channels in the same layer. 

\subsubsection{Results}

Table \ref{tab:cifar10} presents the results of our experiments on CIFAR-$10$.
Using SBN improves the accuracy for all ResNets and VGGs. The additional gain in performance by using SBN in the ResNet50 network ($+1.40\%$) is worth noting. Using two BN layers increases the performance in some cases.
However, the performance gains are small compared to those of SBN.

\begin{table}[tb]
	\caption{Accuracy of experiments on CIFAR-$10$. \label{tab:cifar10}}
	\begin{tabular*}{\columnwidth}{@{\extracolsep{\fill}}lc} 
		\toprule
		\textbf{Model} & \textbf{Accuracy} \tabularnewline
		\midrule
		ResNet18  & $94.04\% \pm 0.08\%$ \tabularnewline
		ResNet18 + 2BN34 & $93.90\% \pm 0.26\%$ \tabularnewline
		ResNet18 + SBN34 & $\mathbf{94.43\% \pm 0.11\%}$ \tabularnewline
		\midrule
		ResNet34  & $93.69\% \pm 0.30\%$ \tabularnewline
		ResNet34 + 2BN34 &$94.01\% \pm 0.22\%$ \tabularnewline
		ResNet34 + SBN34 & $\mathbf{94.67\% \pm 0.18\%}$ \tabularnewline
		\midrule
		ResNet50  & $93.31\% \pm 0.36\%$  \tabularnewline
		ResNet50 + 2BN34 & $93.75\% \pm 0.19\%$ \tabularnewline
		ResNet50 + SBN34 & $\mathbf{94.72\% \pm 0.18\%}$ \tabularnewline
		\midrule
		VGG16-BN & $93.27\% \pm 0.12\%$ \tabularnewline
		VGG16-BN + 2BN4 & $93.29\% \pm 0.15\%$ \tabularnewline
		VGG16-BN + SBN4 & $\mathbf{93.48\% \pm 0.19\%}$ \tabularnewline
		\midrule
		VGG19-BN  & $93.21\% \pm 0.08\%$  \tabularnewline   
		VGG19-BN + 2BN34 & $93.15\% \pm 0.15\%$\tabularnewline
		VGG19-BN + SBN34 & $\mathbf{93.34\% \pm 0.10\%}$  \tabularnewline
		\bottomrule
	\end{tabular*}
\end{table}

Table \ref{tab:cifar100} shows the results of our experiments on CIFAR-$100$.
Similar to the experiments on CIFAR-$10$, ResNets generalize worse if more
parameters are trainable. Additionally, the gain in performance by using SBN in
the ResNet50 network ($+2.31\%$) and VGG19 ($+0.66\%$) are worth noting. The use
of two BN layers slightly increases the performance in some cases. However, 
the performance gains are small compared to those of SBN.

\begin{table}[tb]
	\caption{Accuracy of experiments on CIFAR-$100$. \label{tab:cifar100}}
	\begin{tabular*}{\columnwidth}{@{\extracolsep{\fill}}lc} 
		\toprule
		\textbf{Model} & \textbf{Accuracy} \tabularnewline
		\midrule
		ResNet18 & $76.47\% \pm 0.20\%$ \tabularnewline
		ResNet18 + 2BN34 & $76.41\% \pm 0.15\%$ \tabularnewline
		ResNet18 + SBN34 & $\mathbf{77.11\% \pm 0.30\%}$ \tabularnewline
		\midrule
		ResNet34 & $77.07\% \pm 0.45\%$ \tabularnewline
		ResNet34 + 2BN34 & $77.05\% \pm 0.25\%$\tabularnewline
		ResNet34 + SBN34 & $\mathbf{78.01\% \pm 0.45\%}$ \tabularnewline
		\midrule
		ResNet50 & $76.17\% \pm 0.69\%$ \tabularnewline
		ResNet50 + 2BN34 & $76.53\% \pm 1.03\%$\tabularnewline
		ResNet50 + SBN34 & $\mathbf{78.49\% \pm 0.24\%}$ \tabularnewline
		\midrule
		VGG16-BN  & $72.48\% \pm 0.35\%$ \tabularnewline
		VGG16-BN + 2BN4 &$72.45\% \pm 0.17\%$ \tabularnewline
		VGG16-BN + SN4 & $\mathbf{72.79\% \pm 0.27\%}$ \tabularnewline
		\midrule
		VGG19-BN & $71.34\% \pm 0.14\%$ \tabularnewline
		VGG19-BN + 2BN4 & $71.30\% \pm 0.30 \%$ \tabularnewline
		VGG19-BN + SBN4 & $\mathbf{72.00\% \pm 0.26\%}$ \tabularnewline
		\bottomrule
	\end{tabular*} 
\end{table}

\subsection{Image Classification on TinyImageNet}

TinyImageNet \cite{2015_Le_CONF} contains
100,000 images of 200 classes downsized to $64\times64$
colored images. Each class has 500 training images and 50 validation images.

The experiments were run five times with different random seeds for 300 epochs
using ResNet$18$ and ResNet$50$. Following the common practice, we report the
top-1 classification accuracy on the validation set.

\subsubsection{Implementation Details}

For our experiments we used PyTorch 1.10.1 \cite{2019_Paszke_CONF} and one
Nvidia GeForce 1080Ti GPU. 

We trained all the models using the SGD optimizer with momentum $0.9$ and 
initial learning rate $0.1$ decayed by factor $0.1$ at epochs $75$, $150$, and
$225$. Moreover, we used weight decay with the factor $1e-4$. The batch size is
set to $128$. 

The training data is augmented by using random crop with size $224$, random
horizontal flip and normalization. The validation data is augmented by resizing
to $256\times256$, using center crop with size $224$ and normalization.

\subsubsection{Results}

Table \ref{tab:TinyImageNet} presents the top-1 accuracy of our experiments on
TinyImageNet. ResNet18 using SBN in the third and fourth block outperforms the
standard ResNet18 by $1.16\%$. Using two BN layers in the third and fourth block
slightly improves ResNet$18$ but slightly worsens the performance of ResNet$50$.
The additional gain in performance for ResNet$50$ is worth noting since using SBN improved the baseline by $2.36\%$. 

\begin{table}[tb]
	\caption{Top-1 Accuracy of experiments on TinyImageNet. \label{tab:TinyImageNet}}
	\begin{tabular*}{\columnwidth}{@{\extracolsep{\fill}}lc} 
		\toprule
		\textbf{Model} & \textbf{Accuracy} \tabularnewline
		\midrule
		ResNet18 & $63.52\% \pm 0.25\%$ \tabularnewline
		ResNet18 + 2BN34 & $63.60\% \pm 0.07\%$ \tabularnewline
		ResNet18 + SBN34 & $\mathbf{64.69\% \pm 0.23\%}$ \tabularnewline
		\midrule
		ResNet50 & $65.42\% \pm 0.60\%$ \tabularnewline
		ResNet50 + 2BN4 & $65.35\% \pm 0.26\%$ \tabularnewline
		ResNet50 + SBN4 & $\mathbf{67.79\% \pm 0.22\%}$ \tabularnewline
		\bottomrule
	\end{tabular*} 
\end{table}

\subsection{ImageNet}

The ILSVRC 2012 classification dataset \cite{2015_Russakovsky} contains 1.2
million training images, 50,000 validation images and 150,000 test images.
They are labeled with 1,000 categories. 

Regularization methods on ImageNet require a greater number of training epochs
to converge \cite{2018_Ghiasi_CONF}. The experiments were run three times
with different random seeds for 400 epochs using ResNet$50$. Following the
common practice, we report the top-1 classification accuracy on the validation
set. 

\subsubsection{Implementation Details}

For our experiments we used PyTorch 1.10.1 \cite{2019_Paszke_CONF} and 4
Nvidia GeForce 1080Ti GPU. 

We used the same hyperparameter setting as \cite{2019_Yun_CONF}. We trained all
the models using the SGD optimizer with momentum $0.9$ and  initial learning
rate $0.1$ decayed by factor $0.1$ at epochs $75$, $150$, $225$, $300$ and
$375$. Moreover, we used weight decay with the factor $1e-4$. The batch size is
set to 256.

The training data and validation data is augmented as described for
TinyImageNet.

\subsubsection{Results}

Table \ref{tab:ImageNet} presents the top-1 accuracy of our experiments on
ImageNet. Using SBN in the last block of ResNet50 increases the accuracy by
$0.71\%$ on ImageNet. The usage of two BN layers increases the performance only
by $0.10\%$. This shows that it is advantageous to use SBN.

\begin{table}[tb]
	\caption{Top-1 Accuracy of experiments on ImageNet. \label{tab:ImageNet}}
	\begin{tabular*}{\columnwidth}{@{\extracolsep{\fill}}lc} 
		\toprule
		\textbf{Model} & \textbf{Accuracy} \tabularnewline
		\midrule
		ResNet50 & $76.80\% \pm 0.11\%$ \tabularnewline
		ResNet50 + 2BN4 & $76.90\% \pm 0.08\%$ \tabularnewline
		ResNet50 + SBN4 & $\mathbf{77.51\% \pm 0.07\%}$ \tabularnewline
		\bottomrule
	\end{tabular*} 
\end{table}

\subsection{Effects on the Training}

We used different measures to analyze the impact of SBN on the training of the
networks: (i) we tracked the weight/gradient norm and weight/gradient
distribution for every layer, (ii) we measured the feature map norm and
distribution after every convolution, BN and SBN layer, (iii) we analyzed the
effects of SBN on the magnitude and phase of the frequency components of feature
maps.

We did not see any differences in the weight/gradient norm and weight/gradient
distribution between networks with and without SBN. Moreover, there were no
clear differences in the phases of the frequency components of feature maps
before and after a SBN layer.

However, there were strong differences in the feature map norms and
distributions between DNN with and without SBN. Furthermore, there were large
differences in the magnitudes of the frequency components of feature maps before
and after a SBN layer.

\subsubsection{Differences in the Feature Map Norms and Distributions} \label{differences_feature_map_norm_distribution}

We measured the feature map norm and distribution by looking at the feature map
for an arbitrary batch during training. Figure
\ref{fig:resnet50_cifar100_30_01} shows the feature map norm per epoch of three
networks: (i) a standard ResNet$50$, (ii) a ResNet$50$ using two BN layers and
(iii) a ResNet$50$ using BN and SBN. The output of a SBN layer has a much
smaller feature map norm compared to the output of one or of two BN layers. This
can be seen also in \ref{fig:vgg_cifar100_conv1_2} for the non-residual network
VGG$19$.

\begin{figure*}[tb]
	\centering
	\begin{subfigure}{.5\textwidth}
		\centering
		\includegraphics[width=\linewidth]{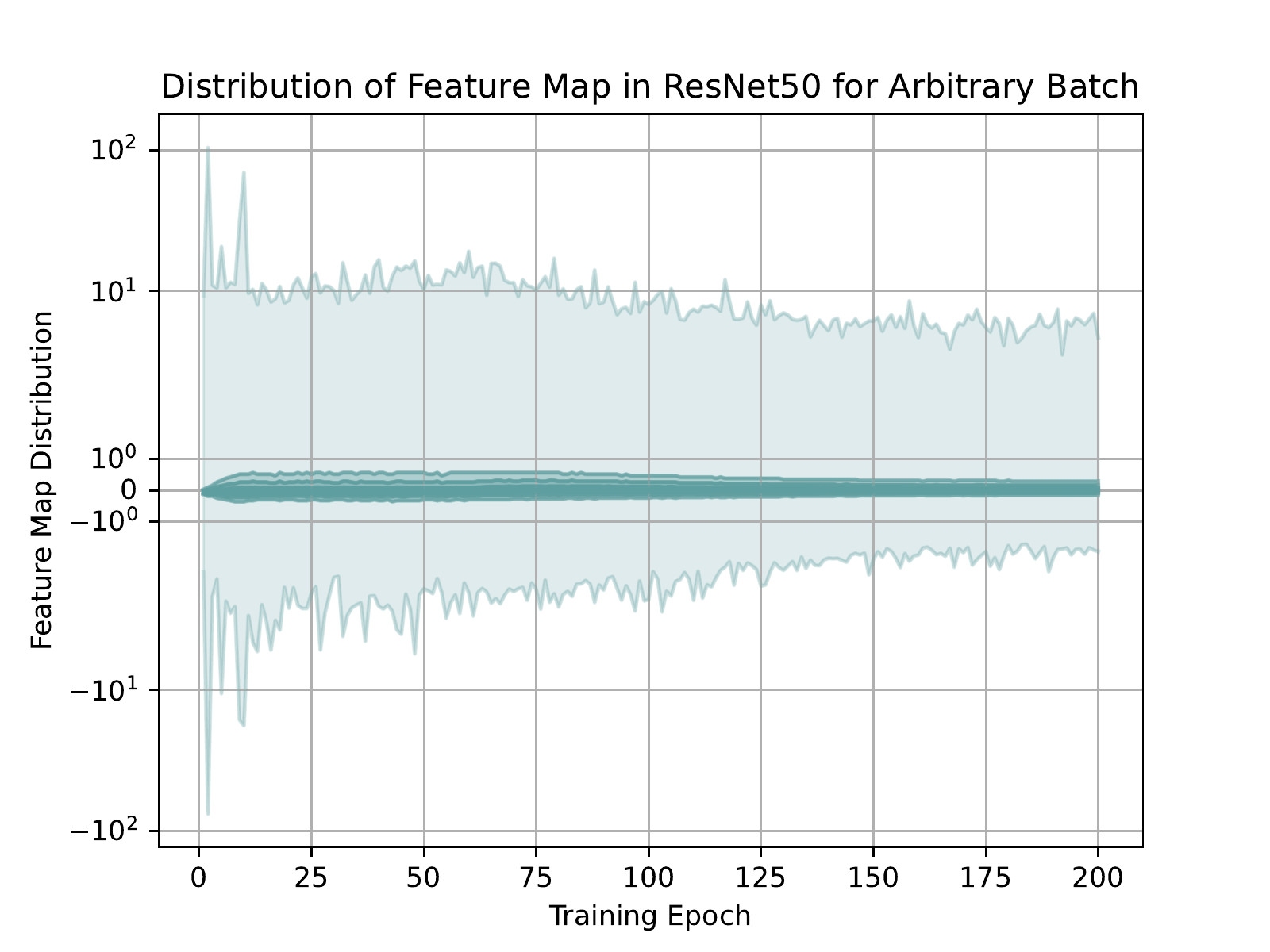}
		\caption{Feature map distribution per epoch after the BN layer.}
		\label{fig:resnet50_fm_l41_bn3}
	\end{subfigure}%
	\begin{subfigure}{.5\textwidth}
		\centering
		\includegraphics[width=\linewidth]{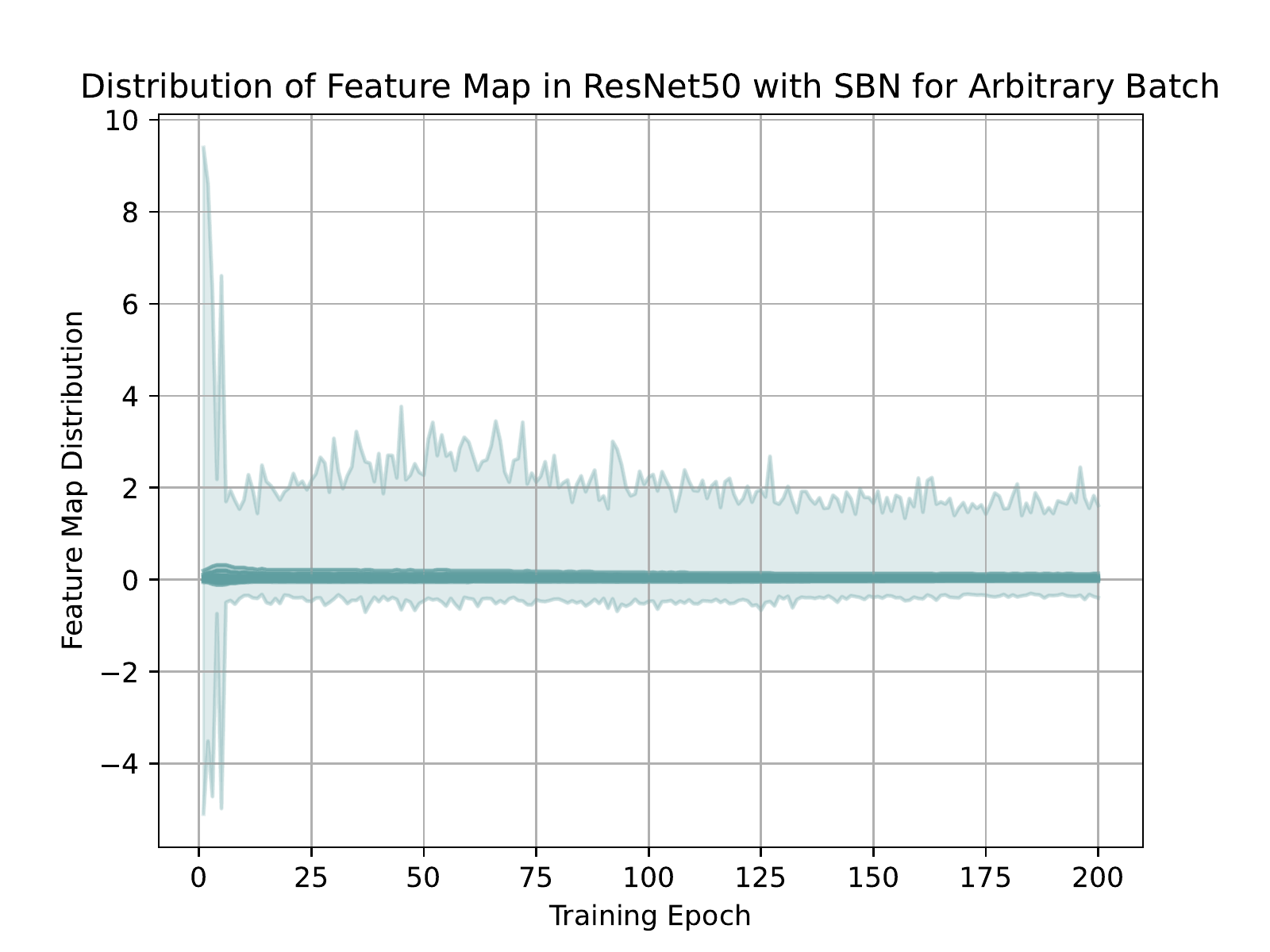}
		\caption{Feature map distribution per epoch after the SBN layer.}
		\label{fig:resnet50_fm_l41_sbn3}
	\end{subfigure}
	\caption{Feature map distribution per epoch after a BN layer
	\ref{fig:resnet50_fm_l41_bn3} of a ResNet$50$ trained on CIFAR$100$ without SBN and after the corresponding SBN layer \ref{fig:resnet50_fm_l41_sbn3} of a ResNet$50$ trained on CIFAR-$100$ with SBN. From top to bottom, the lines represent percentiles with values:
	maximum; 93\%; 84\%; 69\%; 50\%; 31\%; 16\%; 7\%; minimum. Please pay attention on the different scalings in Fig. \ref{fig:resnet50_fm_l41_bn3} and \ref{fig:resnet50_fm_l41_sbn3}.}
\label{fig:feature_map_distribution}
\end{figure*}

Figure \ref{fig:feature_map_distribution} shows the distribution of values in a
feature map after a BN layer of a ResNet$50$ trained on CIFAR-$100$ without SBN
and after the corresponding SBN layer of a ResNet$50$ trained
on CIFAR-$100$ with SBN for an arbitrary batch. Without SBN feature maps have
high values during the whole training. Using SBN prevents large and exploding
values in feature maps. 

\subsubsection{Magnitude of the Frequency Components} \label{effects_spectral_domain}

We analyzed the effects of SBN on the magnitudes of the frequency components of
feature maps. We looked at the feature maps of different images before and after
the SBN layer in fully trained networks. The feature maps were transformed using
the DFT and shifted such that the DC term is moved to the center of the tensor.

\begin{figure*}[tb]
	\centering
	\begin{subfigure}{0.49\linewidth}
		\centering
		\includegraphics[width=\columnwidth]{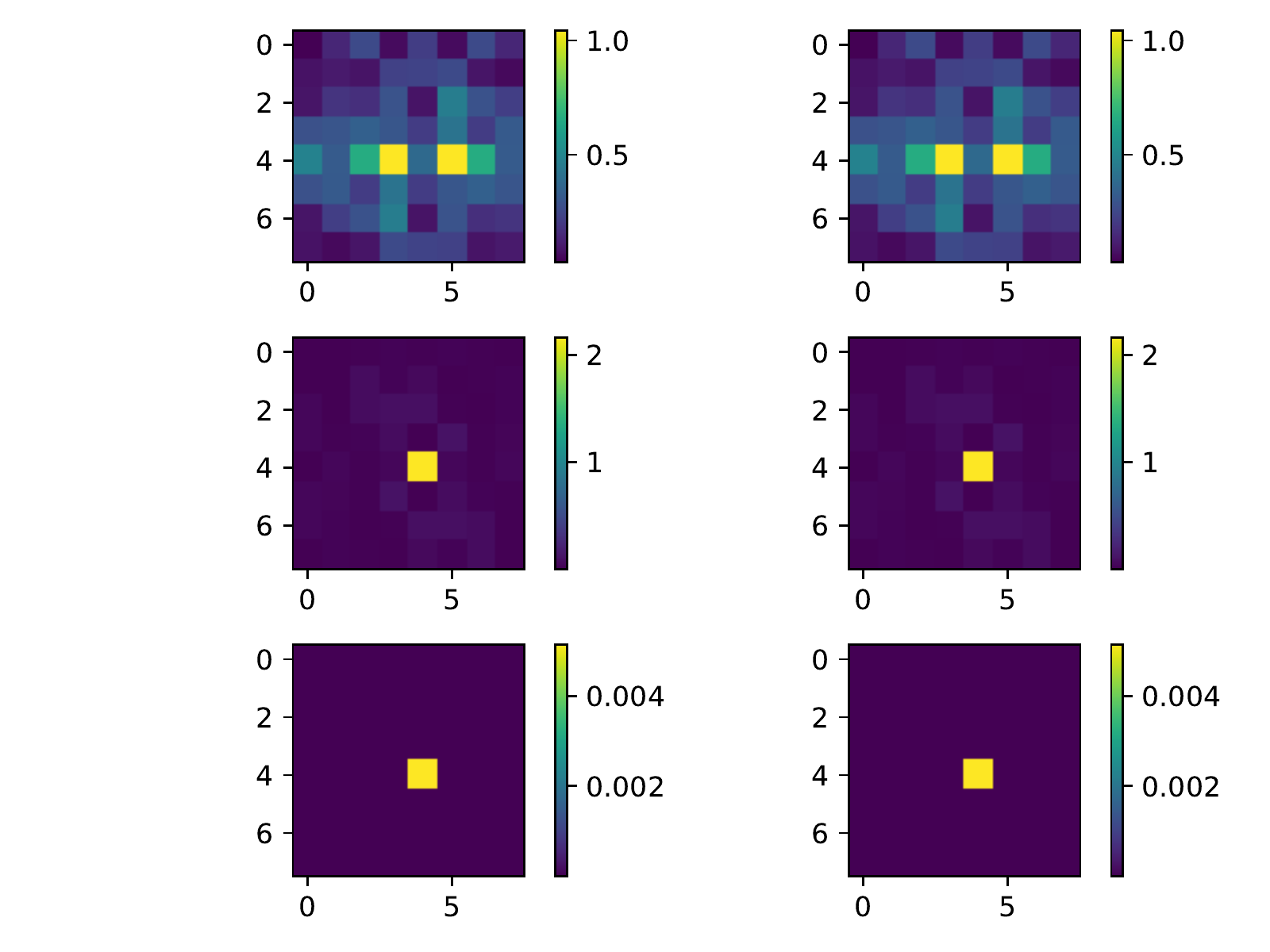}
		\caption{Magnitudes of frequency components of a feature map after a BN layer.}
		\label{fig:resnet50_feature_map_magnitude_bn}
	\end{subfigure}\hfill
	\begin{subfigure}{0.49\linewidth}
		\centering
		\includegraphics[width=\columnwidth]{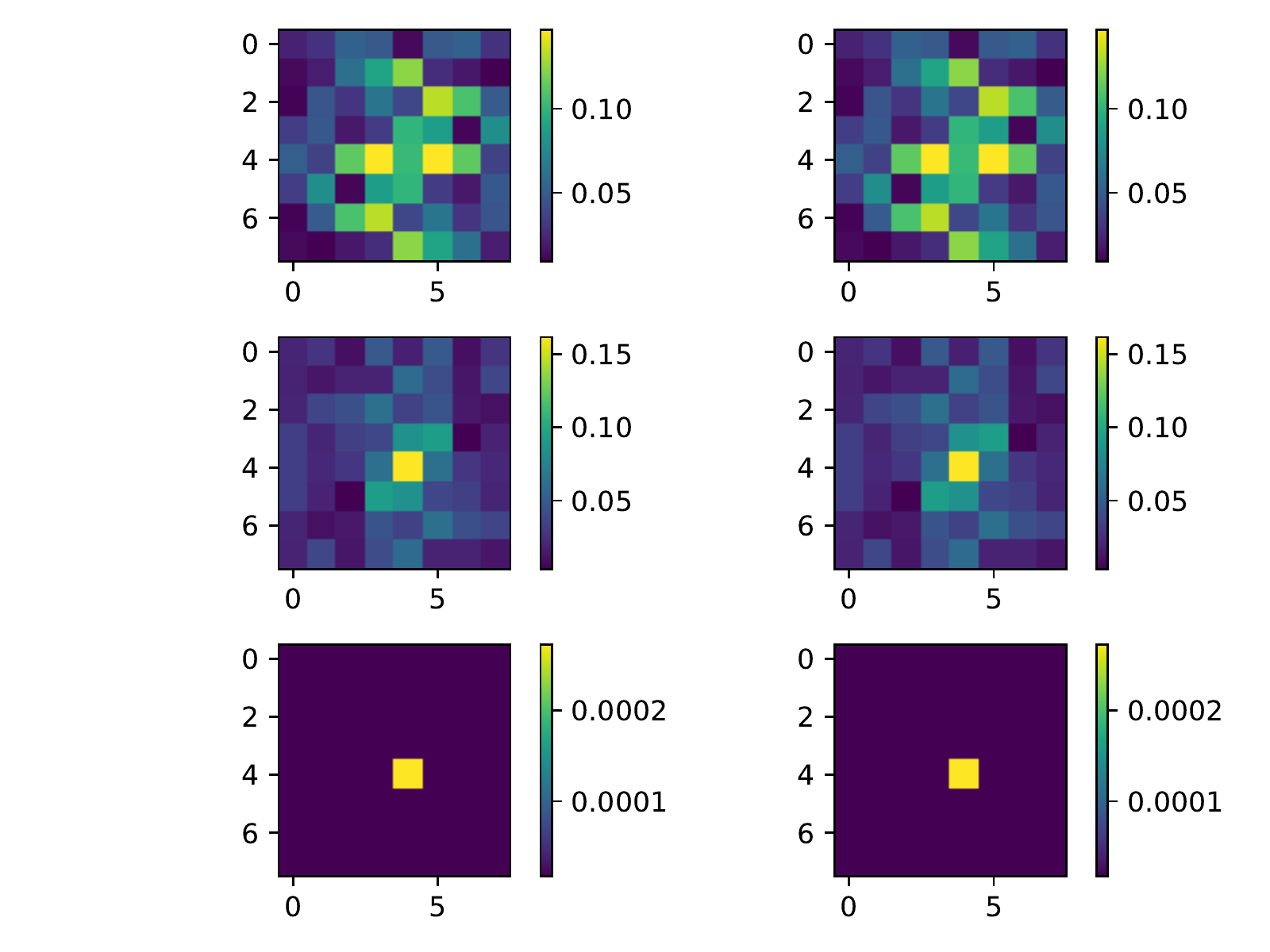}
		\caption{Magnitudes of frequency components of a feature map after a SBN layer.}
		\label{fig:resnet50_feature_map_magnitude_sbn}
	\end{subfigure}
	\caption{Magnitudes of frequency components of different channels of a feature map after a BN layer \ref{fig:resnet50_feature_map_magnitude_bn} and the SBN layer behind it \ref{fig:resnet50_feature_map_magnitude_sbn}.}  \label{fig:feature_map_magnitude}
\end{figure*}

Figure \ref{fig:feature_map_magnitude} shows the magnitudes of frequency
components of different channels in a feature map after a BN layer and after the
SBN layer behind it for one arbitrary image from CIFAR-$100$. It shows
how the magnitude are manipulated due to the SBN layer. SBN reduces the
relevance of frequency components with high magnitudes, i.e. the importance of
frequency components with low magnitude is increased. Hence, the magnitudes of
the frequency components become more uniform distributed. This discourages
reliance on a single frequency component and encourages the network to focus on
more frequency components in the prediction process. Figure
\ref{fig:feature_map_magnitude} generalizes to other images and is not an unique
phenomenon.

\section{Ablation Study / FAQ} \label{ablation_study}

\subsection{Does SBN allow higher learning rates?}

Similar to BN, SBN allows the usage of higher learning rates. We tested
initial learning rates of $0.2$ and $0.5$ with ResNet$50$ on
CIFAR-$10$/$100$ over three runs and could achieve better results using BN and SBN than only using BN (see Table \ref{tab:lr_analysis}). The accuracy of ResNet$50$ with SBN on CIFAR-$100$ using a learning rate of $0.5$ is lower than the baseline, however, the baseline algorithm did not perform well either. 

\begin{table}[tb]
	\caption{Results of learning rate experiments for ResNet50 on CIFAR-$10$/$100$ \label{tab:lr_analysis}.}
	\begin{tabular*}{\columnwidth}{@{\extracolsep{\fill}}lllc} 
		\toprule
		\textbf{Model} & \textbf{LR} & \textbf{Dataset} & \textbf{Accuracy} \tabularnewline
		\midrule
		ResNet50 &  $0.2$ &  CIFAR-$10$/CIFAR-$100$ & $93.69\%/77.19\%$ \tabularnewline
		ResNet50 + SBN34 &  $0.2$ &  CIFAR-$10$/CIFAR-$100$ & $\mathbf{95.08\%/77.91\%}$ \tabularnewline
		ResNet50 &  $0.5$ &  CIFAR-$10$/CIFAR-$100$ & $93.58\%/\mathbf{66.67}\%$ \tabularnewline
		ResNet50 + SBN34 &  $0.5$ &  CIFAR-$10$/CIFAR-$100$ & $\mathbf{94.59}\%/65.82\%$ \tabularnewline
		\bottomrule
	\end{tabular*} 
\end{table}

\subsection{Is there an acceleration of the training? How high is the computational overhead of SBN?}

\begin{figure}[tb]
	\includegraphics[width=\columnwidth]{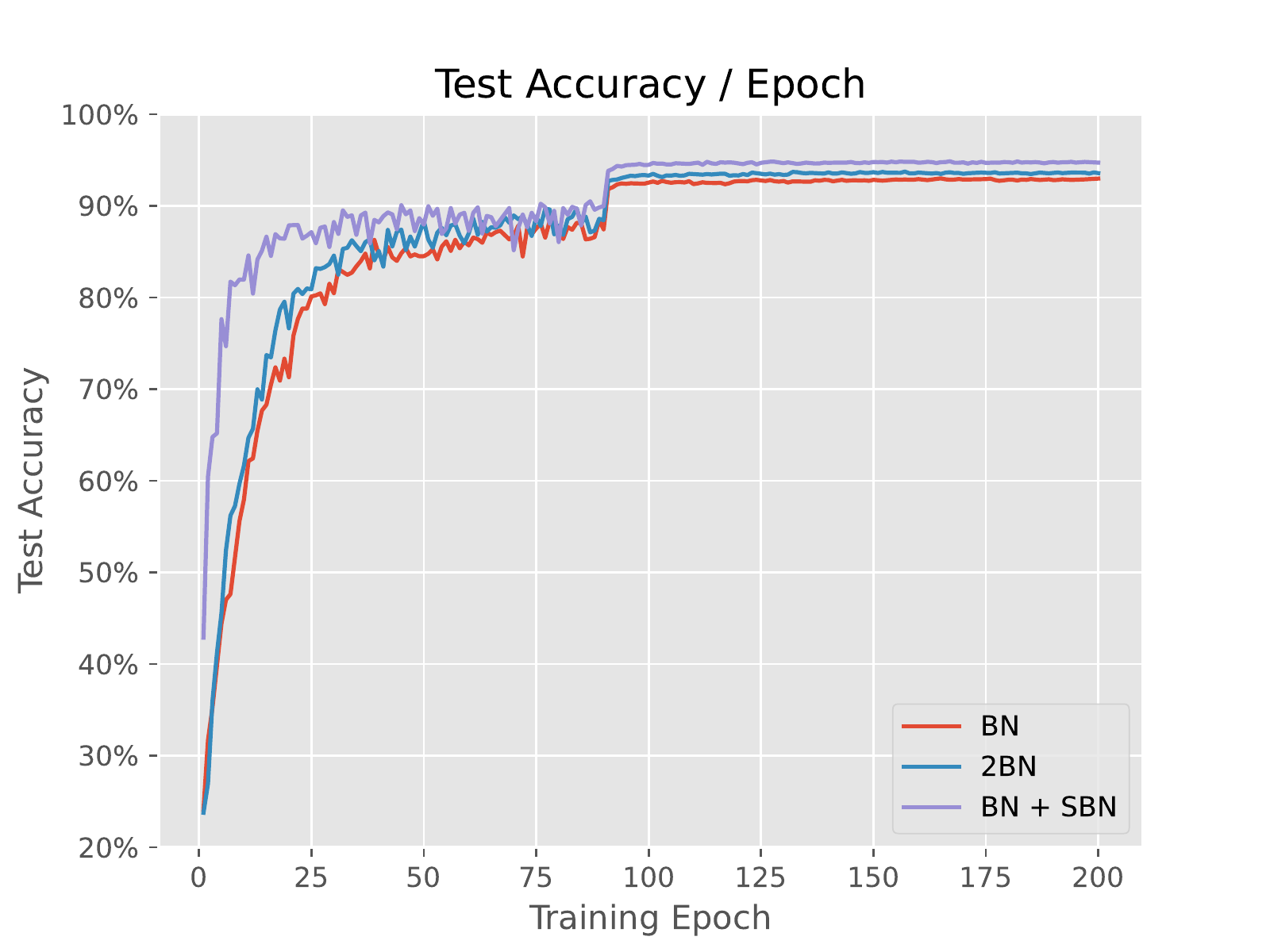}
	\caption{Test Accuracy of ResNet50, ResNet50 with 2BN layer and ResNet50 with SBN on CIFAR-$10$.}
	\label{fig:resnet50_cifar10_acc}
\end{figure}

For easy tasks (e.g. training ResNet$18$/$34$/$50$ on CIFAR-$10$) we saw a huge
acceleration of the training measured per epoch (see Figure
\ref{fig:resnet50_cifar10_acc}). However, due to the Fourier transformations,
the training time is longer compared to the baseline. The training time overhead
of SBN depends on the network architecture. Clearly, if more SBN layers are
used, the training time overhead increases. Table \ref{tab:overhead} shows the
training time overhead of ResNets on CIFAR-$10$/CIFAR-$100$ and ImageNet. Note
that we use the real 2D DFT implementation of PyTorch \cite{2019_Paszke_CONF},
which uses the FFT algorithm to compute the transformation.

\begin{table}[tb]
	\caption{Approximate training time overhead of ResNets on CIFAR$10$/CIFAR$100$ and ImageNet when using SBN in \%. \label{tab:overhead}}
	\begin{tabular*}{\columnwidth}{@{\extracolsep{\fill}}llc} 
		\toprule
		\textbf{Model} & \textbf{Dataset} & \textbf{Training time overhead} \tabularnewline
		\midrule
		ResNet18 & CIFAR-$10$/CIFAR-$100$ & $16\%$/$16\%$ \tabularnewline
		ResNet34 & CIFAR-$10$/CIFAR-$100$ & $23\%$/$27\%$ \tabularnewline
		ResNet50 & CIFAR-$10$/CIFAR-$100$ & $50\%$/$50\%$ \tabularnewline
		ResNet50 & ImageNet & $12\%$ \tabularnewline
		\bottomrule
	\end{tabular*} 
\end{table}

\subsection{Where to insert SBN?} \label{where_SBN}
We performed an ablation study to find the perfect position for inserting a SBN
layer. Firstly, it is preferred to insert a SBN after a BN layer. Inserting SBN
prior to a BN layer does not improve generalization since the BN layer scales
the feature maps up again. Hence, the input feature map of the following
convolutional layer again has a large feature map norm due to the normalization
process of BN. Using a SBN without a BN layer does not improve the
generalization since SBN does not replace BN. The normalization process in the
spatial domain is needed for the training. However, using SBN in addition to BN
increases the performance as shown in the previous experiments.

Beyond that, it is favorable to insert SBN in deeper layers, e.g. in ResNets
for example in modules 3/4. The reason for that is the phenomenon of large
or exploding feature maps happening more often in deeper than in shallow layers
\cite{2019_Zhang_CONFa}. However,
inserting SBN after every BN layer in ResNet18/34/50 did also improve the
accuracy on CIFAR-$10$ and CIFAR-$100$ but we got the best results applying SBN
only in the modules 3/4. When the task gets even harder (e.g. ImageNet), it is
preferred to insert it only in the fourth block. Otherwise, the regularizing
effects are to strong. 

\subsection{Is it sufficient to only subtract the expectation?}
The generalization boost can be partly reproduced by only subtracting the
expectation. However, it does not work as well as applying the whole SBN
algorithm.

\subsection{Is it sufficient to only divide by the standard deviation?}
Same answer as for the question above.

\subsection{Is it sufficient to only normalize the feature map in the frequency
	space without re-scaling and re-shifting?}
Same answer as for the question above.

\subsection{Is it sufficient to only down-scale the feature map in the frequency
	representation?}
The generalization boost can be partly reproduced by downscaling the frequency representation. However, it does not work as well as
applying the whole SBN algorithm. Moreover, this introduces one or more
hyperparameters depending on whether one decides to divide all feature maps by one single value or different feature maps by different values.

\subsection{Is it sufficient to only-down scale the feature map?}
Same answer as for the question above.

\subsection{Is it sufficient to only use weighting of the feature maps in the
	frequency representation?}
The generalization boost cannot be reproduced by only weighting the feature
maps in the frequency domain. However, we need the weighting in the algorithm
since only normalizing the feature map in the frequency domain does not
reproduce the whole generalization boost.

\subsection{Is it favorable to normalize real and imaginary part of the frequency components separately?}
It is not favorable to normalize the real and imaginary part separately as it
has negative effects on the performance. 

\section{Conclusion and Future Work}

We presented spectral batch normalization (SBN), a novel effective method to
improve generalization by normalizing feature maps in the frequency (spectral)
domain. SBN prevents large feature maps, introduces stochastic noise in the
frequency domain of the feature maps and leads to more uniform distributed frequency components. These effects act as regularizers during training.

Using SBN in addition to commonly used regularization methods (e.g.
\cite{2019_Shorten,2015_Ioffe_CONF,1991_Krogh_CONF, 2019_Loshchilov_CONF,
	2014_Srivastava}) increases the performance of ResNets and
VGGs on CIFAR-$10$, CIFAR-$100$ and ImageNet. The additional gain in performance
of ResNet$50$ on CIFAR-$10$ by $1.40\%$, on CIFAR-$100$ by $2.32\%$ and on
ImageNet by $0.71\%$ are worth noting. 

We have not explored the full range of possibilities of SBN. Our
future work will include applying SBN to other normalization techniques, e.g.
Layer Normalization \cite{2016_Ba}, Group Normalization \cite{2020_Wu}, Instance
Normalization \cite{2016_Ulyanov}. Moreover, we will perform further
investigations to analyze the impact on the loss landscape as in
\cite{2018_Santurkar_CONF}. This could be extended to a theoretical analysis of
SBN where we analyze the impact on the Lipschitz continuity of the loss. Furthermore, the weights $\gamma$ and biases $\beta$ could be learned in the complex domain which could be advantageous for training.

\FloatBarrier

\bibliographystyle{IEEEtran}
\bibliography{Spectralnormalization}

\begin{thebibliography}{10}
\providecommand{\url}[1]{#1}
\csname url@samestyle\endcsname
\providecommand{\newblock}{\relax}
\providecommand{\bibinfo}[2]{#2}
\providecommand{\BIBentrySTDinterwordspacing}{\spaceskip=0pt\relax}
\providecommand{\BIBentryALTinterwordstretchfactor}{4}
\providecommand{\BIBentryALTinterwordspacing}{\spaceskip=\fontdimen2\font plus
\BIBentryALTinterwordstretchfactor\fontdimen3\font minus
  \fontdimen4\font\relax}
\providecommand{\BIBforeignlanguage}[2]{{%
\expandafter\ifx\csname l@#1\endcsname\relax
\typeout{** WARNING: IEEEtran.bst: No hyphenation pattern has been}%
\typeout{** loaded for the language `#1'. Using the pattern for}%
\typeout{** the default language instead.}%
\else
\language=\csname l@#1\endcsname
\fi
#2}}
\providecommand{\BIBdecl}{\relax}
\BIBdecl

\bibitem{2014_Srivastava}
\BIBentryALTinterwordspacing
N.~Srivastava, G.~E. Hinton, A.~Krizhevsky, I.~Sutskever, and R.~Salakhutdinov,
  ``Dropout: a simple way to prevent neural networks from overfitting,''
  \emph{J. Mach. Learn. Res.}, vol.~15, no.~1, pp. 1929--1958, 2014. [Online].
  Available: \url{http://dl.acm.org/citation.cfm?id=2670313}
\BIBentrySTDinterwordspacing

\bibitem{2013_Wan_CONF}
\BIBentryALTinterwordspacing
L.~Wan, M.~D. Zeiler, S.~Zhang, Y.~LeCun, and R.~Fergus, ``{R}egularization of
  {N}eural {N}etworks using {D}rop{C}onnect,'' in \emph{Proceedings of the 30th
  International Conference on Machine Learning, {ICML} 2013, Atlanta, GA, USA,
  16-21 June 2013}, ser. {JMLR} Workshop and Conference Proceedings,
  vol.~28.\hskip 1em plus 0.5em minus 0.4em\relax JMLR.org, 2013, pp.
  1058--1066. [Online]. Available:
  \url{http://proceedings.mlr.press/v28/wan13.html}
\BIBentrySTDinterwordspacing

\bibitem{2016_Goodfellow_BOOK}
\BIBentryALTinterwordspacing
I.~J. Goodfellow, Y.~Bengio, and A.~C. Courville, \emph{{D}eep {L}earning},
  ser. Adaptive computation and machine learning.\hskip 1em plus 0.5em minus
  0.4em\relax {MIT} Press, 2016. [Online]. Available:
  \url{http://www.deeplearningbook.org/}
\BIBentrySTDinterwordspacing

\bibitem{1996_Tibshirani}
\BIBentryALTinterwordspacing
R.~Tibshirani, ``{R}egression {S}hrinkage and {S}election via the {L}asso,''
  \emph{Journal of the Royal Statistical Society. Series B (Methodological)},
  vol.~58, no.~1, pp. 267--288, 1996. [Online]. Available:
  \url{http://www.jstor.org/stable/2346178}
\BIBentrySTDinterwordspacing

\bibitem{1991_Krogh_CONF}
\BIBentryALTinterwordspacing
A.~Krogh and J.~A. Hertz, ``{A} {S}imple {W}eight {D}ecay {C}an {I}mprove
  {G}eneralization,'' in \emph{Advances in Neural Information Processing
  Systems 4, {[NIPS} Conference, Denver, Colorado, USA, December 2-5, 1991]},
  J.~E. Moody, S.~J. Hanson, and R.~Lippmann, Eds.\hskip 1em plus 0.5em minus
  0.4em\relax Morgan Kaufmann, 1991, pp. 950--957. [Online]. Available:
  \url{http://papers.nips.cc/paper/563-a-simple-weight-decay-can-improve-generalization}
\BIBentrySTDinterwordspacing

\bibitem{1992_Nowlan}
\BIBentryALTinterwordspacing
S.~J. Nowlan and G.~E. Hinton, ``{S}implifying {N}eural {N}etworks by {S}oft
  {W}eight-{S}haring,'' \emph{Neural Comput.}, vol.~4, no.~4, pp. 473--493,
  1992. [Online]. Available: \url{https://doi.org/10.1162/neco.1992.4.4.473}
\BIBentrySTDinterwordspacing

\bibitem{2019_Shorten}
\BIBentryALTinterwordspacing
C.~Shorten and T.~M. Khoshgoftaar, ``{A} survey on {I}mage {D}ata
  {A}ugmentation for {D}eep {L}earning,'' \emph{J. Big Data}, vol.~6, p.~60,
  2019. [Online]. Available: \url{https://doi.org/10.1186/s40537-019-0197-0}
\BIBentrySTDinterwordspacing

\bibitem{1999_Opitz}
\BIBentryALTinterwordspacing
D.~W. Opitz and R.~Maclin, ``{P}opular {E}nsemble {M}ethods: {A}n {E}mpirical
  {S}tudy,'' \emph{J. Artif. Intell. Res.}, vol.~11, pp. 169--198, 1999.
  [Online]. Available: \url{https://doi.org/10.1613/jair.614}
\BIBentrySTDinterwordspacing

\bibitem{2015_Ioffe_CONF}
\BIBentryALTinterwordspacing
S.~Ioffe and C.~Szegedy, ``{B}atch {N}ormalization: {A}ccelerating {D}eep
  {N}etwork {T}raining by {R}educing {I}nternal {C}ovariate {S}hift,'' in
  \emph{Proceedings of the 32nd International Conference on Machine Learning,
  {ICML} 2015, Lille, France, 6-11 July 2015}, ser. {JMLR} Workshop and
  Conference Proceedings, F.~R. Bach and D.~M. Blei, Eds., vol.~37.\hskip 1em
  plus 0.5em minus 0.4em\relax JMLR.org, 2015, pp. 448--456. [Online].
  Available: \url{http://proceedings.mlr.press/v37/ioffe15.html}
\BIBentrySTDinterwordspacing

\bibitem{2016_Ba}
\BIBentryALTinterwordspacing
L.~J. Ba, J.~R. Kiros, and G.~E. Hinton, ``{L}ayer {N}ormalization,''
  \emph{CoRR}, vol. abs/1607.06450, 2016. [Online]. Available:
  \url{http://arxiv.org/abs/1607.06450}
\BIBentrySTDinterwordspacing

\bibitem{2020_Wu}
\BIBentryALTinterwordspacing
Y.~Wu and K.~He, ``{G}roup {N}ormalization,'' \emph{Int. J. Comput. Vis.}, vol.
  128, no.~3, pp. 742--755, 2020. [Online]. Available:
  \url{https://doi.org/10.1007/s11263-019-01198-w}
\BIBentrySTDinterwordspacing

\bibitem{2016_Ulyanov}
\BIBentryALTinterwordspacing
D.~Ulyanov, A.~Vedaldi, and V.~S. Lempitsky, ``{I}nstance {N}ormalization:
  {T}he {M}issing {I}ngredient for {F}ast {S}tylization,'' \emph{CoRR}, vol.
  abs/1607.08022, 2016. [Online]. Available:
  \url{http://arxiv.org/abs/1607.08022}
\BIBentrySTDinterwordspacing

\bibitem{2021_Dauphin_CONF}
\BIBentryALTinterwordspacing
Y.~Dauphin and E.~D. Cubuk, ``{D}econstructing the {R}egularization of
  {B}atch{N}orm,'' in \emph{9th International Conference on Learning
  Representations, {ICLR} 2021, Virtual Event, Austria, May 3-7, 2021}.\hskip
  1em plus 0.5em minus 0.4em\relax OpenReview.net, 2021. [Online]. Available:
  \url{https://openreview.net/forum?id=d-XzF81Wg1}
\BIBentrySTDinterwordspacing

\bibitem{2019_Luo_CONF}
\BIBentryALTinterwordspacing
P.~Luo, X.~Wang, W.~Shao, and Z.~Peng, ``{T}owards {U}nderstanding
  {R}egularization in {B}atch {N}ormalization,'' in \emph{7th International
  Conference on Learning Representations, {ICLR} 2019, New Orleans, LA, USA,
  May 6-9, 2019}.\hskip 1em plus 0.5em minus 0.4em\relax OpenReview.net, 2019.
  [Online]. Available: \url{https://openreview.net/forum?id=HJlLKjR9FQ}
\BIBentrySTDinterwordspacing

\bibitem{2019_Zhang_CONFa}
\BIBentryALTinterwordspacing
H.~Zhang, Y.~N. Dauphin, and T.~Ma, ``{F}ixup {I}nitialization: {R}esidual
  {L}earning {W}ithout {N}ormalization,'' in \emph{7th International Conference
  on Learning Representations, {ICLR} 2019, New Orleans, LA, USA, May 6-9,
  2019}.\hskip 1em plus 0.5em minus 0.4em\relax OpenReview.net, 2019. [Online].
  Available: \url{https://openreview.net/forum?id=H1gsz30cKX}
\BIBentrySTDinterwordspacing

\bibitem{2017_Kukacka}
\BIBentryALTinterwordspacing
J.~Kukacka, V.~Golkov, and D.~Cremers, ``{R}egularization for {D}eep
  {L}earning: {A} {T}axonomy,'' \emph{CoRR}, vol. abs/1710.10686, 2017.
  [Online]. Available: \url{http://arxiv.org/abs/1710.10686}
\BIBentrySTDinterwordspacing

\bibitem{2017_DeVries}
\BIBentryALTinterwordspacing
T.~Devries and G.~W. Taylor, ``{I}mproved {R}egularization of {C}onvolutional
  {N}eural {N}etworks with {C}utout,'' \emph{CoRR}, vol. abs/1708.04552, 2017.
  [Online]. Available: \url{http://arxiv.org/abs/1708.04552}
\BIBentrySTDinterwordspacing

\bibitem{2018_Ghiasi_CONF}
\BIBentryALTinterwordspacing
G.~Ghiasi, T.-Y. Lin, and Q.~V. Le, ``{D}rop{B}lock: {A} regularization method
  for convolutional networks,'' in \emph{Advances in Neural Information
  Processing Systems}, S.~Bengio, H.~Wallach, H.~Larochelle, K.~Grauman,
  N.~Cesa-Bianchi, and R.~Garnett, Eds., vol.~31.\hskip 1em plus 0.5em minus
  0.4em\relax Curran Associates, Inc., 2018. [Online]. Available:
  \url{https://proceedings.neurips.cc/paper/2018/file/7edcfb2d8f6a659ef4cd1e6c9b6d7079-Paper.pdf}
\BIBentrySTDinterwordspacing

\bibitem{2019_Loshchilov_CONF}
\BIBentryALTinterwordspacing
I.~Loshchilov and F.~Hutter, ``{D}ecoupled {W}eight {D}ecay {R}egularization,''
  in \emph{7th International Conference on Learning Representations, {ICLR}
  2019, New Orleans, LA, USA, May 6-9, 2019}.\hskip 1em plus 0.5em minus
  0.4em\relax OpenReview.net, 2019. [Online]. Available:
  \url{https://openreview.net/forum?id=Bkg6RiCqY7}
\BIBentrySTDinterwordspacing

\bibitem{Rumelhart1987}
D.~E. Rumelhart and J.~L. McClelland, \emph{{L}earning {I}nternal
  {R}epresentations by {E}rror {P}ropagation}, 1987, pp. 318--362.

\bibitem{2016_Salimans_CONF}
\BIBentryALTinterwordspacing
T.~Salimans and D.~P. Kingma, ``{W}eight {N}ormalization: {A} {S}imple
  {R}eparameterization to {A}ccelerate {T}raining of {D}eep {N}eural
  {N}etworks,'' in \emph{Advances in Neural Information Processing Systems 29:
  Annual Conference on Neural Information Processing Systems 2016, December
  5-10, 2016, Barcelona, Spain}, D.~D. Lee, M.~Sugiyama, U.~von Luxburg,
  I.~Guyon, and R.~Garnett, Eds., 2016, p. 901. [Online]. Available:
  \url{https://proceedings.neurips.cc/paper/2016/hash/ed265bc903a5a097f61d3ec064d96d2e-Abstract.html}
\BIBentrySTDinterwordspacing

\bibitem{2018_Santurkar_CONF}
\BIBentryALTinterwordspacing
S.~Santurkar, D.~Tsipras, A.~Ilyas, and A.~Madry, ``{H}ow {D}oes {B}atch
  {N}ormalization {H}elp {O}ptimization?'' in \emph{Advances in Neural
  Information Processing Systems 31: Annual Conference on Neural Information
  Processing Systems 2018, NeurIPS 2018, December 3-8, 2018, Montr{\'{e}}al,
  Canada}, S.~Bengio, H.~M. Wallach, H.~Larochelle, K.~Grauman,
  N.~Cesa{-}Bianchi, and R.~Garnett, Eds., 2018, pp. 2488--2498. [Online].
  Available:
  \url{https://proceedings.neurips.cc/paper/2018/hash/905056c1ac1dad141560467e0a99e1cf-Abstract.html}
\BIBentrySTDinterwordspacing

\bibitem{2015_Rippel_CONF}
\BIBentryALTinterwordspacing
O.~Rippel, J.~Snoek, and R.~P. Adams, ``{S}pectral {R}epresentations for
  {C}onvolutional {N}eural {N}etworks,'' in \emph{Advances in Neural
  Information Processing Systems 28: Annual Conference on Neural Information
  Processing Systems 2015, December 7-12, 2015, Montreal, Quebec, Canada},
  C.~Cortes, N.~D. Lawrence, D.~D. Lee, M.~Sugiyama, and R.~Garnett, Eds.,
  2015, pp. 2449--2457. [Online]. Available:
  \url{https://proceedings.neurips.cc/paper/2015/hash/536a76f94cf7535158f66cfbd4b113b6-Abstract.html}
\BIBentrySTDinterwordspacing

\bibitem{2019_Khan}
\BIBentryALTinterwordspacing
S.~H. Khan, M.~Hayat, and F.~Porikli, ``Regularization of {D}eep {N}eural
  {N}etworks with {S}pectral {D}ropout,'' \emph{Neural Networks}, vol. 110, pp.
  82--90, 2019. [Online]. Available:
  \url{https://doi.org/10.1016/j.neunet.2018.09.009}
\BIBentrySTDinterwordspacing

\bibitem{2018_Liu_CONF}
\BIBentryALTinterwordspacing
Z.~Liu, J.~Xu, X.~Peng, and R.~Xiong, ``{F}requency-{D}omain {D}ynamic
  {P}runing for {C}onvolutional {N}eural {N}etworks,'' in \emph{Advances in
  Neural Information Processing Systems 31: Annual Conference on Neural
  Information Processing Systems 2018, NeurIPS 2018, December 3-8, 2018,
  Montr{\'{e}}al, Canada}, S.~Bengio, H.~M. Wallach, H.~Larochelle, K.~Grauman,
  N.~Cesa{-}Bianchi, and R.~Garnett, Eds., 2018, pp. 1051--1061. [Online].
  Available:
  \url{https://proceedings.neurips.cc/paper/2018/hash/a9a6653e48976138166de32772b1bf40-Abstract.html}
\BIBentrySTDinterwordspacing

\bibitem{2021_Bassey}
\BIBentryALTinterwordspacing
J.~Bassey, L.~Qian, and X.~Li, ``{A} {S}urvey of {C}omplex-{V}alued {N}eural
  networks,'' \emph{CoRR}, vol. abs/2101.12249, 2021. [Online]. Available:
  \url{https://arxiv.org/abs/2101.12249}
\BIBentrySTDinterwordspacing

\bibitem{2019_Paszke_CONF}
\BIBentryALTinterwordspacing
A.~Paszke, S.~Gross, F.~Massa, A.~Lerer, J.~Bradbury, G.~Chanan, T.~Killeen,
  Z.~Lin, N.~Gimelshein, L.~Antiga, A.~Desmaison, A.~K{\"{o}}pf, E.~Z. Yang,
  Z.~DeVito, M.~Raison, A.~Tejani, S.~Chilamkurthy, B.~Steiner, L.~Fang,
  J.~Bai, and S.~Chintala, ``{P}y{T}orch: {A}n {I}mperative {S}tyle,
  {H}igh-{P}erformance {D}eep {L}earning {L}ibrary,'' in \emph{Advances in
  Neural Information Processing Systems 32: Annual Conference on Neural
  Information Processing Systems 2019, NeurIPS 2019, December 8-14, 2019,
  Vancouver, BC, Canada}, H.~M. Wallach, H.~Larochelle, A.~Beygelzimer,
  F.~d'Alch{\'{e}}{-}Buc, E.~B. Fox, and R.~Garnett, Eds., 2019, pp.
  8024--8035. [Online]. Available:
  \url{https://proceedings.neurips.cc/paper/2019/hash/bdbca288fee7f92f2bfa9f7012727740-Abstract.html}
\BIBentrySTDinterwordspacing

\bibitem{2009_Krizhevsky}
A.~Krizhevsky and G.~Hinton, ``Learning multiple layers of features from tiny
  images,'' \emph{Master's thesis, Department of Computer Science, University
  of Toronto}, 2009.

\bibitem{2016_He_CONF}
\BIBentryALTinterwordspacing
K.~He, X.~Zhang, S.~Ren, and J.~Sun, ``{D}eep {R}esidual {L}earning for {I}mage
  {R}ecognition,'' in \emph{2016 {IEEE} Conference on Computer Vision and
  Pattern Recognition, {CVPR} 2016, Las Vegas, NV, USA, June 27-30,
  2016}.\hskip 1em plus 0.5em minus 0.4em\relax {IEEE} Computer Society, 2016,
  pp. 770--778. [Online]. Available: \url{https://doi.org/10.1109/CVPR.2016.90}
\BIBentrySTDinterwordspacing

\bibitem{2015_Simonyan_CONF}
\BIBentryALTinterwordspacing
K.~Simonyan and A.~Zisserman, ``{V}ery {D}eep {C}onvolutional {N}etworks for
  {L}arge-{S}cale {I}mage {R}ecognition,'' in \emph{3rd International
  Conference on Learning Representations, {ICLR} 2015, San Diego, CA, USA, May
  7-9, 2015, Conference Track Proceedings}, Y.~Bengio and Y.~LeCun, Eds., 2015.
  [Online]. Available: \url{http://arxiv.org/abs/1409.1556}
\BIBentrySTDinterwordspacing

\bibitem{2016_Zagoruyko_CONF}
\BIBentryALTinterwordspacing
S.~Zagoruyko and N.~Komodakis, ``{W}ide {R}esidual {N}etworks,'' in
  \emph{Proceedings of the British Machine Vision Conference (BMVC)}, E.~R.~H.
  Richard C.~Wilson and W.~A.~P. Smith, Eds.\hskip 1em plus 0.5em minus
  0.4em\relax BMVA Press, September 2016, pp. 87.1--87.12. [Online]. Available:
  \url{https://dx.doi.org/10.5244/C.30.87}
\BIBentrySTDinterwordspacing

\bibitem{2015_He_CONF}
\BIBentryALTinterwordspacing
K.~He, X.~Zhang, S.~Ren, and J.~Sun, ``{D}elving {D}eep into {R}ectifiers:
  {S}urpassing {H}uman-{L}evel {P}erformance on {I}mage{N}et
  {C}lassification,'' in \emph{2015 {IEEE} International Conference on Computer
  Vision, {ICCV} 2015, Santiago, Chile, December 7-13, 2015}.\hskip 1em plus
  0.5em minus 0.4em\relax {IEEE} Computer Society, 2015, pp. 1026--1034.
  [Online]. Available: \url{https://doi.org/10.1109/ICCV.2015.123}
\BIBentrySTDinterwordspacing

\bibitem{2015_Le_CONF}
Y.~Le and X.~S. Yang, ``{T}iny {I}mage{N}et {V}isual {R}ecognition
  {C}hallenge,'' 2015.

\bibitem{2015_Russakovsky}
\BIBentryALTinterwordspacing
O.~Russakovsky, J.~Deng, H.~Su, J.~Krause, S.~Satheesh, S.~Ma, Z.~Huang,
  A.~Karpathy, A.~Khosla, M.~S. Bernstein, A.~C. Berg, and L.~Fei{-}Fei,
  ``{I}mage{N}et {L}arge {S}cale {V}isual {R}ecognition {C}hallenge,''
  \emph{Int. J. Comput. Vis.}, vol. 115, no.~3, pp. 211--252, 2015. [Online].
  Available: \url{https://doi.org/10.1007/s11263-015-0816-y}
\BIBentrySTDinterwordspacing

\bibitem{2019_Yun_CONF}
\BIBentryALTinterwordspacing
S.~Yun, D.~Han, S.~Chun, S.~J. Oh, Y.~Yoo, and J.~Choe, ``{C}ut{M}ix:
  {R}egularization {S}trategy to {T}rain {S}trong {C}lassifiers {W}ith
  {L}ocalizable {F}eatures,'' in \emph{2019 {IEEE/CVF} International Conference
  on Computer Vision, {ICCV} 2019, Seoul, Korea (South), October 27 - November
  2, 2019}.\hskip 1em plus 0.5em minus 0.4em\relax {IEEE}, 2019, pp.
  6022--6031. [Online]. Available:
  \url{https://doi.org/10.1109/ICCV.2019.00612}
\BIBentrySTDinterwordspacing

\end{thebibliography}

\end{document}